\documentclass[final]{cvpr}

\usepackage{times}
\usepackage{epsfig}
\usepackage{graphicx}
\usepackage{amsmath}
\usepackage{amssymb}

\usepackage{booktabs}
\usepackage{tabularx}

\newcommand{\ve}[1]{\mathbf{#1}}

\newcommand{\tabincell}[2]{\begin{tabular}{@{}#1@{}}#2\end{tabular}}  

\usepackage[pagebackref=true,breaklinks=true,colorlinks,bookmarks=false]{hyperref}

\def\cvprPaperID{838} %
\def\confYear{CVPR 2021}

\begin{document}

\title{Understanding of Kernels in CNN Models by \\ Suppressing Irrelevant Visual Features in Images}

\author{Jia-Xin Zhuang\\
Sun Yat-sen University\\
{\tt\small lincolnz9511@gmail.com}
\and
Wanying Tao\\
Sun Yat-sen University\\
{\tt\small taowy3@mail2.sysu.edu.cn}
\and 
Jianfei Xing\\
Sun Yat-sen University\\
{\tt\small xingjf3@mail2.sysu.edu.cn}\\
\and
Wei Shi\\
Sun Yat-sen University\\
{\tt\small fuermowei@foxmail.com}
\and 
Ruixuan Wang\\
Sun Yat-sen University\\
{\tt\small wangruix5@mail.sysu.edu.cn}
\and
Wei-Shi Zheng\\
Sun Yat-sen University\\
{\tt\small wszheng@ieee.org}
}

\maketitle

\begin{abstract}
Deep learning models have shown their superior performance in various vision tasks. However, the lack of precisely interpreting kernels in convolutional neural networks (CNNs) is becoming one main obstacle to wide applications of deep learning models in real scenarios. Although existing interpretation methods may find certain visual patterns which are associated with the activation of a specific kernel, those visual patterns may not be specific or comprehensive enough for interpretation of a specific activation of kernel of interest. In this paper, a simple yet effective optimization method is proposed to interpret the activation of any kernel of interest in CNN models. The basic idea is to simultaneously preserve the activation of the specific kernel and suppress the activation of all other kernels at the same layer. In this way, only visual information relevant to the activation of the specific kernel is remained in the input. Consistent visual information from multiple modified inputs would help users understand what kind of features are specifically associated with specific kernel. 
Comprehensive evaluation shows that the proposed method can help better interpret activation of specific kernels than widely used methods, even when two kernels have very similar activation regions from the same input image.

\end{abstract}

\section{Introduction}

Convolutional Neural Networks (CNNs) have shown human-level performance in various computer vision tasks and have started to be applied in real scenarios, such as surveillance~\cite{ye2020deep}, face identification~\cite{masi2018deep}, autonomous driving~\cite{li2019gs3d}, and medical image analysis~\cite{litjens2017survey,shen2017deep}. However, CNN models are still lack of interpretation, in the sense that users are not clear what visual information a CNN model uses for a specific decision given an input to the model, or what visual features are associated with or would activate a specific convolutional kernel in a CNN model. The very limited interpretation of CNN models has become one main obstacle to the trust of CNN decisions or predictions particularly in high-risk applications like medical diagnosis and autonomous driving.  

	\begin{figure}[!tb]
		\centering
		\includegraphics[width=0.45\textwidth]{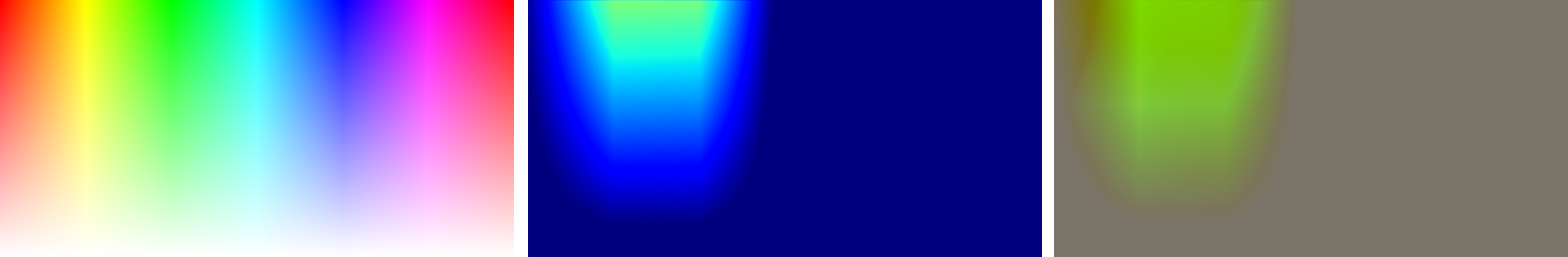}
		\caption{Interpreting the activation of one specific kernel in a pre-trained CNN model. The original input color image (left) contains almost all the colors in the RGB space. For the response of the 16-th kernel in the first convolutional layer (middle), the optimized input (right) by our method suggests that the kernel is activated purely by the green color information, although the activated regions in the original input contain some other color information.}
    	\label{fig:many_color}
	\end{figure}

CNN interpretation has started to be investigated mainly in the task of classification, where a CNN classifier is trained in advance based on a multi-class dataset. In order to interpret model decisions, several approaches have been recently proposed mainly by estimating the contribution of input image regions or pixels to a specific (often maximum) output value of the CNN model. Ideas include perturbing certain image regions and checking the change of the CNN output~\cite{petsiuk2018rise,ribeiro2016should,zeiler2014visualizing,zhou2014learning}, finding the relevant activated regions based on the output (feature maps) of the last convolutional layer~\cite{chattopadhay2018grad,selvaraju2017grad,zhou2016learning}, directly establishing a simple approximate linear model to estimate the linear relationship between local image regions and the CNN output~\cite{ribeiro2016should}, or back-propagating the CNN output layer-by-layer to the input image space either based on gradient information (or its modified versions) at each layer or based on the relevance between input elements and output at each layer~\cite{bach2015pixel,montavon2017explaining}. Besides the above ideas, directly maximizing the (pre-softmax) output of a specific neuron at the final output layer with respect to the changeable input image can also give users the general impression of what the neuron represents~\cite{mahendran2015understanding,mordvintsev2015inceptionism,simonyan2013deep} based on the observation of the optimized input image. 
The back-propagation ideas and the maximization idea can also be applied to interpretation of any internal neuron (convolutional kernel at a convolutional layer or output element at a fully connected layer) of the CNN model, where the obtained visual patterns in the image space can be considered as representation or interpretation of the neuron. In addition, the consistency between manually segmented regions of certain concept (or class) and the activated regions in the output of a convolutional kernel can help people interpret the kernel by the semantic concept~\cite{bau2017network,bau2018gan}. All the approaches to interpretation of internal or output neurons focus on finding specific visual patterns/concepts or image regions relevant to the neurons. However, there could exist many more (if not unlimited number of) visual patterns activating a specific neuron, and maybe just part of features in a region is relevant to the activation of the neuron for a specific input image. Therefore, the existing approaches may not comprehensively and precisely interpret neurons.

In this study, we propose a simple yet effective method to interpret particularly convolutional kernels by trying to preserve the visual features in the image space which contributes to the activation of a specific kernel, and to suppress the other visual features which is irrelevant to the activation of the specific kernel, where some irrelevant features may come from the same image region as that of the relevant features. Different from existing approaches which aim to reduce or maximize activation of a specific kernel, our method aims to keep the activation. Also different from approaches which aim to find visual patterns often partly contributing to the specific activation of a kernel given any input image, our method aims to find the visual features in the original image which contribute to the exact activation of the kernel (Figure~\ref{fig:many_color}).

\section{Related Work}
Interpretation of deep learning models has been recently explored mainly with the objective of understanding output prediction and internal neurons of CNN models. 

\noindent\textbf{Interpretation of model predictions only:} Multiple approaches have been proposed only for understanding of model's prediction output, including the perturbation approach~\cite{petsiuk2018rise,zeiler2014visualizing,zhou2014object}, the activation map approach~\cite{chattopadhay2018grad,selvaraju2017grad,zhou2016learning}, the local approximation approach~\cite{fong2017interpretable,zhou2014object}, and the prototype-based approach~\cite{kim2016examples,van2019interpretable}. The perturbation approach perturbs local image regions somehow and checks the change of a specific CNN output. If certain perturbed local region(s) causes large drop in the output, the visual information in the local region in the original input image is crucial and therefore used to explain the original prediction output.
Perturbation can be in the form of simply masking a local region by a constant pixel intensity~\cite{zeiler2014visualizing}, by neighboring image patches~\cite{zintgraf2017visualizing}, or by fusing the original region information with constant or blurring~\cite{binder2016layer,kindermans2017learning,petsiuk2018rise}. A mask could be small and only cover an image patch each time~\cite{petsiuk2018rise}. It also could be of the image size and the degree of masking at each pixel is automatically estimated~\cite{ribeiro2016should}, often with the constraints that the to-be-masked area is small and continuous, and the masking should be smooth~\cite{fong2019understanding,van2019interpretable}. The perturbation approach can often find the approximate local image regions relevant to the specific model prediction output, but it is not clear whether all the visual information in the local region equally contributes to the prediction or not. 

Different from the perturbation approach exploring regions in the input image, the activation map approach tries to find relevant region(s) in the output feature maps often from the last convolutional layer~\cite{chattopadhay2018grad,selvaraju2017grad,wang2020score,zhou2016learning}. By estimating the weight of each feature map either directly from the model parameters (in the following fully connected layer)~\cite{zhou2016learning} or from the gradient of the specific model output with respect to the feature map~\cite{selvaraju2017grad}, a class-specific (or output-specific) activation map can be obtained with the weighted sum of all feature maps at the convolutional layer. Stronger activation region in the activation map indicates that the corresponding region in the input image is responsible for the specific model output. Since the size of the activation map is often much smaller than that of the input image, the interpolation of the activation map up to the size of the input image would often cause very approximate localization of the image region(s) contributing to the model output prediction. 

Similar to the perturbation approach without requiring knowing model parameters, the local approximation approach provides another way to estimate the contribution of each meaning image region (e.g., object parts, similar background region)~\cite{bach2015pixel,binder2016layer}. It assumes that the model decision surface is locally linear in the feature space, where each feature component could correspond to an image region. By approximating the original model with a linear model in the feature space, the contribution of each meaningful image region to the model output can be directly obtained from the weight parameters in the linear model. This approach needs an optimization process to find a unique linear model for each input image. Together with the activation map approach, this approach shares the same issue as that of the perturbation approach, i.e., considering all visual information in certain region relevant to the model prediction.

Different from all the above three approaches, the prototype-based approach simulates human's behavior when trying to interpret a decision, i.e., based on the similarity between the feature of new data and typical features of each class~\cite{van2019interpretable}. The prototypes (typical feature representations and associated representative image patches) for each class and the integral similarity measurement between a new image and prototypes can be automatically learned based on a unified deep convolutional model~\cite{kim2016examples}. One issue of this approach is about the determination of appropriate number of prototypes for each class. 

\noindent\textbf{Interpretation of both internal and output neurons:} The back-propagation approach has been widely used to interpret both internal and output neurons of CNN models~\cite{baehrens2010explain,binder2016layer,li2018patternnet,simonyan2013deep,zeiler2014visualizing}.
Given an input image and the output of a specific neuron, the simplest back-propagation method is to calculate the gradient of the neuron output with respect to each input pixel, and then the quantity of the gradient at each pixel is considered as the importance of the pixel for the activation of the neuron output~\cite{baehrens2010explain,simonyan2013deep}. 
Each gradient element could be conditionally propagated backward particularly around the commonly used ReLU activation function, e.g., only if the propagated gradient element is positive or if both the gradient and the forward signal are positive~\cite{springenberg2014striving,zeiler2014visualizing}, resulting in different levels of sparsity in the final gradient-based visual pattern in the input image space. However, considering that the original gradient element at each CNN layer is from the model weight parameters which are not correlated with the input information, the overall gradient of a neuron output with respect to the input image may not correctly represent the importance of input image pixels to the specific neuron output~\cite{baehrens2010explain,simonyan2013deep,zeiler2014visualizing}. In this case, gradient at each layer could be replaced by input-relevant signals during back-propagation from the neuron output to the input image space~\cite{li2018patternnet}. The importance of each input image pixel could also be estimated not based on gradient but based on the layer-by-layer relevance back-propagation, where the relevance between the output and each input element at one layer can be easily calculated based on the percent contribution of the input element to the output~\cite{binder2016layer}. The resulting visual patterns from the image pixel importance based on each of the back-propagation methods can help users understand the representation of a specific neuron, but the visual patterns may only represent part of the visual signals relevant to the neuron output and do not tell users what exact visual information in the input image contributes to the specific activation of the neuron. The same issue exists in an earlier maximization approach which can obtain rough input visual patterns by maximizing the output of a specific neuron~\cite{mahendran2015understanding,masi2018deep,mordvintsev2015inceptionism}.

\section{Methodology}
\label{sec:methodology}

The purpose of this study is to understand each convolutional kernel (filter) in a pre-trained CNN classifier, in particular what visual features in the image space are associated with or would activate a specific kernel. Given an input image to the CNN classifier and a specific kernel at certain layer, suppose the output (i.e., the feature map) of the kernel is locally activated, it is easy to locate the image region(s) associated with the activated region(s) in the feature map simply by resizing the feature map to the scale of the input image. However, it would be not rigorous if one claims that the visual features in the activated image region cause the specific activation in the kernel's feature map. This is because there often exist multiple features (e.g., color, texture, shape, etc.) in one local image region, and it could be just one type of feature that cause the activation in the feature map, while the other features in the same region could be associated with the activation in the feature maps of some other kernels. Therefore, in order to precisely understand what exact feature(s) in the image activates the specific kernel, it is necessary to disentangle the contributions of these features in the same region to activation of multiple possible kernels.
Here we propose a simple but effective method to find the visual features in the image space which is associated with any specific kernel, by suppressing the visual features which is irrelevant to the activation of the specific kernel but may be relevant to the activation of other kernels, 
 and through our method, we can get multiple results that cause the same activation on the selected kernel.
	
Formally, suppose there are totally $N$ kernels (and therefore correspondingly $N$ feature maps) at the convolutions layer of interest, and the $i$-th kernel $\ve{k}_i$ of this layer is the specific one to be understood. Given an input image $\ve{X}$, denote the output of the kernel by the feature map $\ve{F}_i$ whose size is $H \times W$. In general, some local regions in $\ve{F}_i$ would be activated with values in $\ve{F}_i$ clearly larger than zero, while the other regions would be inactivated with corresponding values close to or being zero in $\ve{F}_i$. To understand what possible visual features in the input image $\ve{X}$ cause the activation in $\ve{F}_i$, we would like to find a modified input image  $\hat{\ve{X}}$ which is expected to contain only the visual features contributing to the activation in $\ve{F}_i$. Denote by $\hat{\ve{F}}_i$ the feature map output of the kernel $\ve{k}_i$ when using $\hat{\ve{X}}$ as input to the CNN model. Then, we expect the difference between the new feature map $\hat{\ve{F}}_i$ and the original feature map $\ve{F}_i$ as small as possible, i.e., minimizing the preservation loss $\mathcal{L}_p(\hat{\ve{X}},\ve{X})$,
\begin{equation}\label{eq:Lp}
	\mathcal{L}_p(\hat{\ve{X}},\ve{X}) = \frac{1}{W \cdot H \cdot \| \ve{F}_i \|_2} \| \hat{\ve{F}}_i - \ve{F}_i\|_2  \,.
\end{equation}
The $L_2$ norm is used to measure the difference because it is more sensitive to large difference than $L_1$ norm. The $\| \ve{F}_i \|_2$ in the denominator is to reduce the sensitivity of certain hyper-parameters (e.g., learning rate during optimization) to varying activation levels at different convolutional layers. Meanwhile, in order to effectively remove all irrelevant features to the activation of the specific kernel $\ve{k}_i$, we assume that every kind of irrelevant feature is associated with one or more other kernels in the same convolutional layer. Base on such an assumption, removing the irrelevant features could be achieved by finding the modified input $\hat{\ve{X}}$ such that the activation of all the other kernels in the layer is suppressed as much as possible when using $\hat{\ve{X}}$ as input to the CNN model, i.e., minimizing the suppression loss
$\mathcal{L}_s(\hat{\ve{X}},\ve{X})$,
\begin{equation}\label{eq:Ls}
	\mathcal{L}_s(\hat{\ve{X}},\ve{X}) = \frac{1}{ W \cdot H \cdot \sum_{j=1, j \neq i}^N \| \ve{F}_j \|_1} \sum_{j=1, j \neq i}^N \| \hat{\ve{F}}_j\|_1  \,.
\end{equation}
The $L_1$ norm is used here to allow large values existing sparsely in some feature maps $\hat{\ve{F}}_j$'s particular when the visual features associated with the specific kernel $\hat{\ve{F}}_i$ are also contributing to the activation of some other kernels. The $\sum_{j=1, j \neq i}^N \| \ve{F}_j \|_1$ in the denominator again is to reduce the sensitivity of certain hyper-parameters to varying activation levels at different convolutional layers.

In addition, considering that various noise signals could be well filtered away through multiple layers of convolutions, it becomes difficult to find a reasonably good solution $\hat{\ve{X}}$ without noise signals because many other similar solutions with small noise signals included could lead to similar feature map activation at the convolutional layer of interest. To effectively remove such noise signals in the final solution, as used in related work~\cite{dabkowski2017real,du2018towards,fong2017interpretable,yosinski2015understanding}, the regularization loss $\mathcal{L}_r(\hat{\ve{X}})$ can be employed,
\begin{equation}
	\mathcal{L}_r(\hat{\ve{X}}) = \frac{1}{W_0 \cdot H_0} \| \hat{\ve{X}}\|_1  \,,
\end{equation}
where $W_0$ and $H_0$ are the width and height of the input $\hat{\ve{X}}$. $L_1$ norm is used because certain non-zero visual features are expected in $\hat{\ve{X}}$ for the specific activation of kernel $\ve{k}_i$.

Overall, given the pre-trained CNN model and the input image $\ve{X}$, the modified input image $\hat{\ve{X}}$ which contains only the necessary visual features for the specific activation $\ve{F}_i$ of the kernel $\ve{k}_i$ could be found by minimizing the combined loss $\mathcal{L}(\hat{\ve{X}},\ve{X})$,
\begin{equation} \label{eq:loss}
	\mathcal{L}(\hat{\ve{X}},\ve{X}) = \mathcal{L}_p(\hat{\ve{X}},\ve{X}) + \alpha \mathcal{L}_s(\hat{\ve{X}},\ve{X}) + \beta \mathcal{L}_r(\hat{\ve{X}})\,,
\end{equation}
where $\alpha$ and $\beta$ are the weight coefficients to trade off the three terms. Once the $\hat{\ve{X}}$ is obtained, the visual features in the original input image $\ve{X}$ contributing to the specific activation of the kernel $\ve{k}_i$ could be directly observed from $\hat{\ve{X}}$. By observing multiple solutions $\hat{\ve{X}}$'s based on multiple original input images $\ve{X}$'s, any consistent visual features in these $\hat{\ve{X}}$'s would help users understand what kind of features are associated with the kernel $\ve{k}_i$.

\section{Experiments}

\subsection{Experimental setup}
The well-known 16-layer \textit{VGG}~\cite{simonyan2013deep} classifier pre-trained on the ImageNet dataset~\cite{ILSVRC15} was used to evaluate our proposed method in kernel interpretation. 
Totally six classes of test images were used as inputs for interpretation of kernels in the CNN classifier. Each image was resized to $224\times224$ pixels. For each input image, given the selected specific kernel $\ve{k}_i$ at certain convolutional layer and the fixed model parameters, gradient descent method was used to optimize the input, with initial learning rate 10 and decayed by 0.5 after training loss stops falling for about 50 iterations. Each optimization was observed convergent within 10,000 iterations. The coefficients $\beta=1.0$ and $\gamma=1.0$ during optimization.
The results were evaluated both qualitatively and quantitatively. The qualitative evaluation is based on the direct observation of the optimized input $\hat{\ve{X}}$'s and the visualization of the feature maps $\ve{F}_i$'s and $\hat{\ve{F}}_i$'s.
Quantitatively, to measure the similarity between the feature map $\hat{\ve{F}}_i$ of the kernel $\ve{k}_i$ for the optimized input image $\hat{\ve{X}}$ and the original feature map $\ve{F}_i$, the structural similarity index (\textit{SSIM})~\cite{wang2004image} is adopted. Mean square error (\textit{MSE}) between all other kernels' feature maps $\hat{\ve{F}}_j$'s and the expected suppression result (i.e., a zero-value feature map) was introduced to evaluate the degree of suppression for all the other kernels' feature maps at the same convolutional layer, with smaller MSE corresponding to better suppression.

\subsection{Effect of suppression}

\begin{figure}[!bth]
		\centering
		\includegraphics[width=0.46\textwidth]{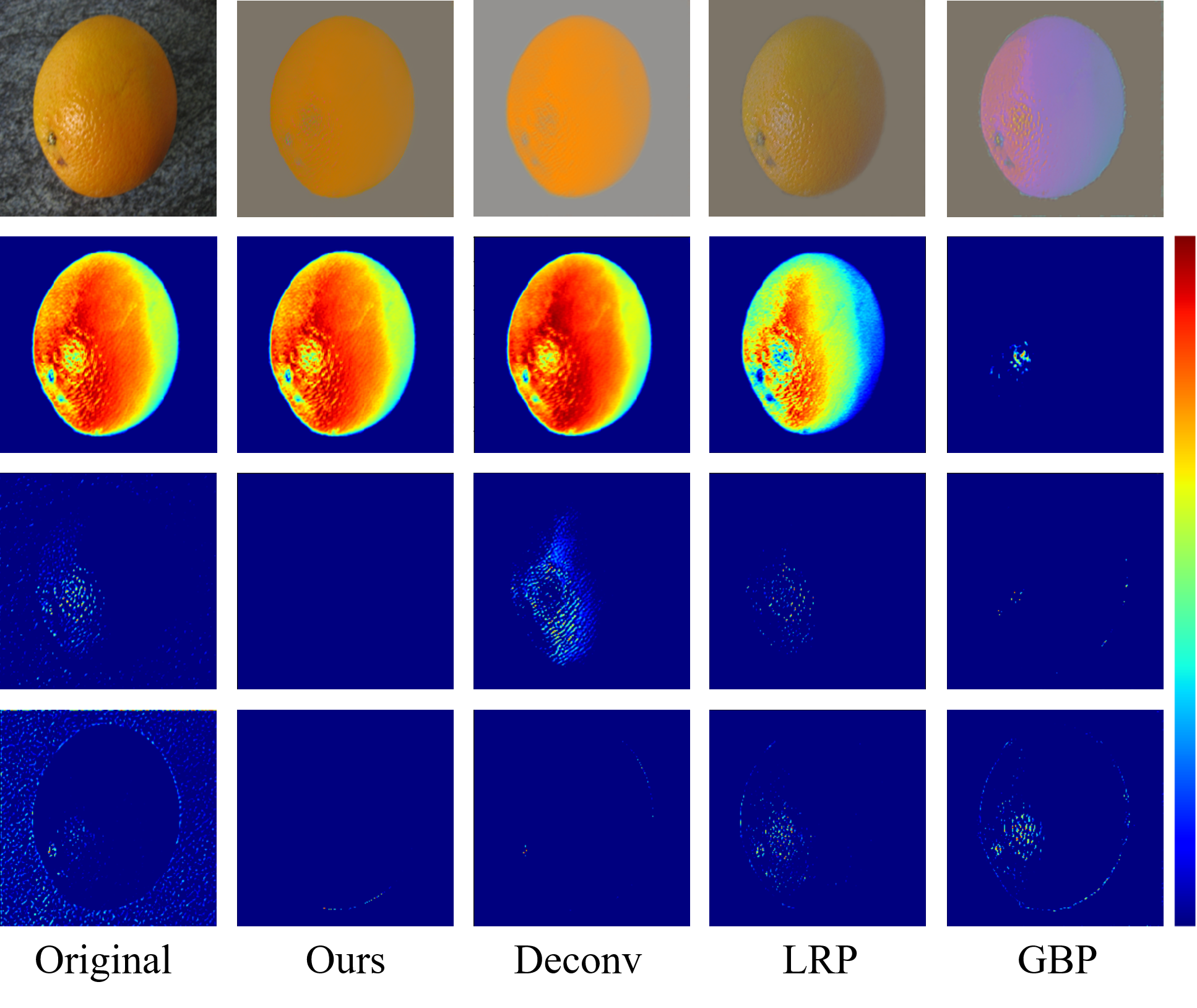}
		\caption{Interpreting the activation of the 20-th kernel at the second convolutional layer given the input image. First row: the original input image and the optimized/estimated inputs by our method and baselines. Second to fourth row: feature map activation of the 20-th kernel and two other representative kernels with the corresponding inputs. Our method can well preserve the activation of the selected kernel and suppress the activation of others.}
    	\label{fig:suppresion}
	\end{figure}

We first check whether the proposed method works as expected or not, i.e., given any original input image, whether the method can find an optimized input such that the activation of a selected specific kernel can be preserved while the activation of all the kernels at the same convolutional layer is suppressed. Some existing methods can also interpret the activation of specific kernels, including the backpropagation-based  Deconvolution (Deconv)~\cite{zeiler2014visualizing}, the guided-backpropagation (GBP)~\cite{springenberg2014striving}, and the layer-wise relevance propagation (LRP)~\cite{binder2016layer,ioffe2015batch}. Therefore, these methods were used as baselines for comparison with ours.
From the demonstrative example in Figure~\ref{fig:suppresion}, it is clear that our method can find a modified input (first row, second image) which cause almost the same activation of the selected kernel (second row, second feature map) as the original activation (second row, first feature map) and meanwhile well suppress the activation of other kernels (second column, bottom two feature maps). In comparison, the estimated inputs from the baseline methods can neither keep the activation of the specific kernel nor well suppress the activation of other kernels (last three columns). This suggests that the estimated inputs from the baselines actually may not precisely interpret the activation of the specific kernel, while our method probably can find a modified input for the specific interpretation of the selected kernel. 

    \begin{table*}[!tbh]
        \centering
        \scriptsize
    	\caption{Quantitative comparison between our method and baselines. For each block in the VGG16 classifier, four kernels were randomly selected from the last convolutional layer in the block, and 180 images were randomly selected from the ImageNet training set for each selected specific kernel. Mean and standard deviation of the SSIM and MSE was calculated over the kernels and input images for the last layer of each block.
    Note that SSIM was for selected kernels (S.) and MSE for all the other kernels (R.).}
    	\label{tab:baselineQuantitative}
    		 \vspace{0.5em}	
    		 \begin{tabularx}{0.95\textwidth}{ccccccccccc}
    		 	\toprule
    		 \multicolumn{1}{c}{Methods} &
    		 \multicolumn{2}{c}{Block 1} &
    		 \multicolumn{2}{c}{Block 2} &
    		 \multicolumn{2}{c}{Block 3} &
    		 \multicolumn{2}{c}{Block 4} &
    		 \multicolumn{2}{c}{Block 5} \\
    		 \cmidrule(l){2-3} \cmidrule(l){4-5} \cmidrule(l){6-7} \cmidrule(l){8-9} \cmidrule(l){10-11}
    		  & SSIM(S.)$\uparrow$ & MSE(R.)$\downarrow$ & SSIM(S.)$\uparrow$ & MSE(R.)$\downarrow$ & SSIM(S.)$\uparrow$ & MSE(R.)$\downarrow$ & SSIM(S.)$\uparrow$ & MSE(R.)$\downarrow$ & SSIM(S.)$\uparrow$ & MSE(R.)$\downarrow$ \\
    		 \midrule
    		 GBP~\cite{springenberg2014striving} &
    		   \tabincell{c}{$0.21$\\ \tiny{$\pm0.29$}} & 
    		   \tabincell{c}{$0.67$\\ \tiny{$\pm0.62$}} &
    		   \tabincell{c}{$0.14$\\ \tiny{$\pm0.18$}} & 
    		   \tabincell{c}{$1.51$\\ \tiny{$\pm0.61$}} &
    		   \tabincell{c}{$0.22$\\ \tiny{$\pm0.15$}} & 
    		   \tabincell{c}{$2.88$\\ \tiny{$\pm1.03$}} &
    		   \tabincell{c}{$0.19$\\ \tiny{$\pm0.14$}} & 
    		   \tabincell{c}{$1.67$\\ \tiny{$\pm1.23$}} &
    		   \tabincell{c}{$0.14$\\ \tiny{$\pm0.18$}} & 
    		   \tabincell{c}{$0.39$\\ \tiny{$\pm0.33$}} \\
    		 Deconv~\cite{zeiler2014visualizing} &
    		   \tabincell{c}{$0.58$\\ \tiny{$\pm0.22$}} & 
    		   \tabincell{c}{$0.54$\\ \tiny{$\pm0.57$}} &
    		   \tabincell{c}{$0.54$\\ \tiny{$\pm0.18$}} & 
    		   \tabincell{c}{$2.24$\\ \tiny{$\pm1.33$}} &
    		   \tabincell{c}{$0.45$\\ \tiny{$\pm0.20$}} & 
    		   \tabincell{c}{$4.25$\\ \tiny{$\pm1.85$}} &
    		   \tabincell{c}{$0.20$\\ \tiny{$\pm0.17$}} & 
    		   \tabincell{c}{$2.10$\\ \tiny{$\pm1.51$}} &
    		   \tabincell{c}{$0.11$\\ \tiny{$\pm0.19$}} & 
    		   \tabincell{c}{$0.39$\\ \tiny{$\pm0.28$}} \\
    		LRP~\cite{binder2016layer} &
    		   \tabincell{c}{$0.43$\\ \tiny{$\pm0.22$}} & 
    		   \tabincell{c}{$\boldsymbol{0.16}$\\ \tiny{$\pm0.16$}} &
    		   \tabincell{c}{$0.18$\\ \tiny{$\pm0.14$}} & 
    		   \tabincell{c}{$0.56$\\ \tiny{$\pm0.37$}} &
    		   \tabincell{c}{$0.14$\\ \tiny{$\pm0.11$}} & 
    		   \tabincell{c}{$1.51$\\ \tiny{$\pm0.96$}} &
    		   \tabincell{c}{$0.17$\\ \tiny{$\pm0.14$}} & 
    		   \tabincell{c}{$1.02$\\ \tiny{$\pm0.96$}} &
    		   \tabincell{c}{$0.14$\\ \tiny{$\pm0.18$}} & 
    		   \tabincell{c}{$0.27$\\ \tiny{$\pm0.32$}} \\
    		Ours &
    		   \tabincell{c}{$\boldsymbol{ 0.90}$\\ \tiny{$\pm0.13$}} & 
    		   \tabincell{c}{$0.25$\\ \tiny{$\pm0.20$}} & 
    		   \tabincell{c}{$\boldsymbol{ 0.92}$\\ \tiny{$\pm0.07$}} & 
    		   \tabincell{c}{$\boldsymbol{ 0.55}$\\ \tiny{$\pm0.03$}} & 
    		   \tabincell{c}{$\boldsymbol{ 0.83}$\\ \tiny{$\pm0.14$}} & 
    		   \tabincell{c}{$\boldsymbol{ 0.34}$\\ \tiny{$\pm0.52$}} & 
    		   \tabincell{c}{$\boldsymbol{ 0.77}$\\ \tiny{$\pm0.17$}} & 
    		   \tabincell{c}{$\boldsymbol{ 0.07}$\\ \tiny{$\pm0.46$}} & 
    		   \tabincell{c}{$\boldsymbol{ 0.79}$\\ \tiny{$\pm0.25$}} & 
    		   \tabincell{c}{$\boldsymbol{ 0.07}$\\ \tiny{$\pm0.46$}} 
    		   \\
    		    \bottomrule
    	    \end{tabularx}
    \end{table*}

Consistent results were obtained with comprehensive and quantitative comparisons. As Table~\ref{tab:baselineQuantitative} shows, for any convolutional block, our method can get much larger SSIM on selected kernels and smaller MSE  (except in Block 1) on the other kernels, which means that our method can well find visual information which contributes to the activation of specific kernels but not of others.
Interestingly, the MSE is much smaller (only 0.07) at the last two blocks, while it is relatively large at the mid-level blocks (i.e., 0.55 and 0.34 for Blocks 2 and 3 respectively). This indicates that different kernels at higher layers probably correspond to different visual information such that the method can find a modified input to suppress activation of the other kernels without affecting the activation of the selected kernel. It also indicates that kernels at middle layers are more likely interrelated such that it is difficult to completely suppress the activation of some other kernels when trying to preserve the activation of the selected kernel. Such finding is consistent with existing study~\cite{bau2017network,morcos2018importance}

    \begin{figure}[!tbh]
	\centering
	\includegraphics[width=0.48\textwidth]{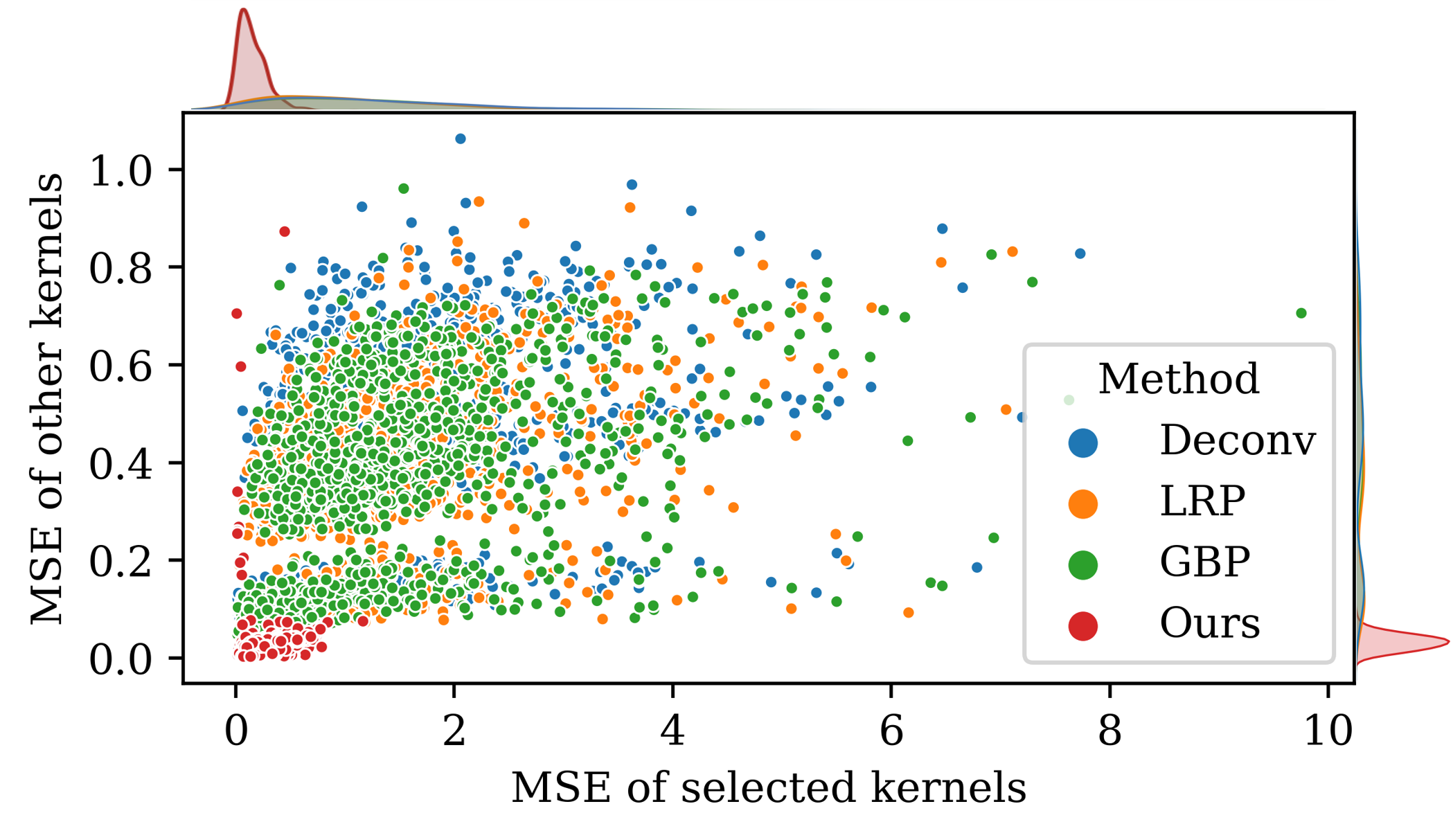}
	\caption{Distributions of MSE's (for selected and other kernels inside Block 4) based on the optimized inputs from each method.
	}
	\label{fig:quan_block4}
\end{figure}

To get a more intuitive perception on the difference between our method and others, a joint distribution plot was drawn as shown in Figure~\ref{fig:quan_block4}, with x-axis representing the  MSE on the activation of the selected kernel and y-axis the MSE on other kernels. Note that for the selected kernel, the MSE is the difference between the response of the kernel to the optimized input and that to the original input.
Each point in the plot corresponds to the paired MSE's for one test image from one method.
On the top and the right side of the joint plot, two marginal distributions of MSE's from each method were plotted respectively. As we can see, our method can get more concentrated and close-to-origin MSE's on both the selected kernels and all other kernels, while other baseline methods result in more spread distributions, again supporting that our method can find optimized inputs which contribute more specifically to the activation of selected kernels.

	\begin{figure*}[tb]
		\centering
        \includegraphics[width=1.0\textwidth]{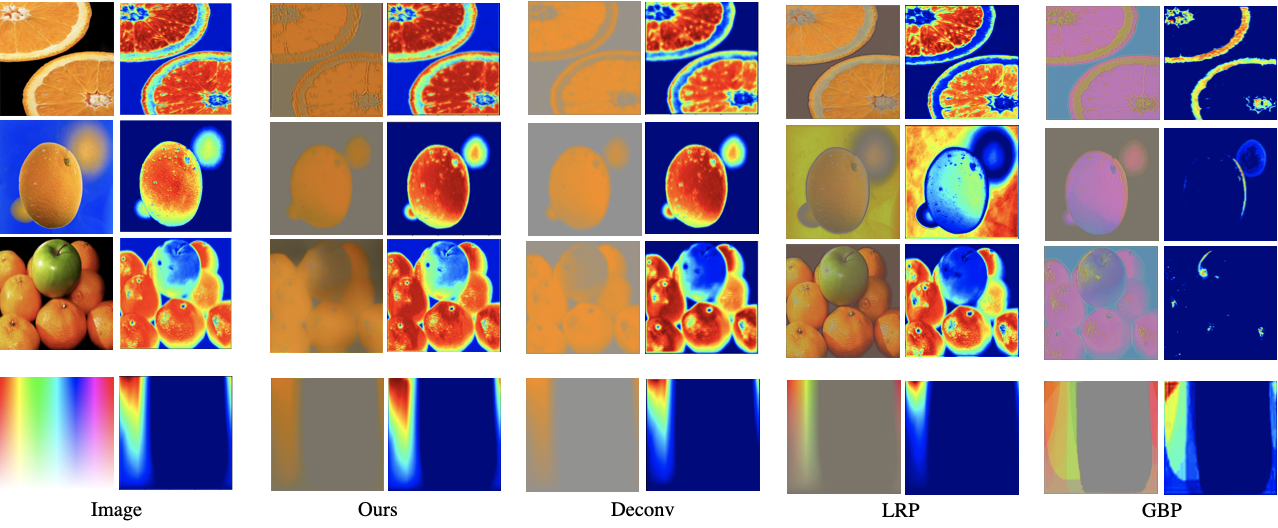}
		\caption{Interpretation of the activation of a specific kernel by different methods. Columns 1-2: original input images and corresponding feature map activation of the kernel. Columns 3-4: optimized inputs and corresponding activation of the kernel by our method. Columns 5-10: optimized inputs and corresponding activation of the kernel, respectively by the baseline Deconv, LRP, and GBP.}
    	\label{fig:baseline_color}
	\end{figure*}

\subsection{Interpretation of kernel activations}
Considering that there is no ground truth for what a specific kernel represents or responds to, one way to evaluate interpretation methods is to visualize and check the consistency in the interpretation results with multiple input instances. As demonstrated in Figure~\ref{fig:baseline_color} (first three rows), for the same kernel, our method and the baseline Deconv consistently found that the orange color causes the specific activation of the kernel, but LRP found inconsistent and more color information and GBP found a completely different color (purple) for the activation. Since the specific kernel is probably responding to color information, one synthetic color image (last row, first) containing all color information was additionally used as input. The activation of the kernel (last row, second) for the synthetic color image tells us that the kernel responds to image regions from red, orange, to yellow colors. Again, both our method and Deconv found the orange information within the activated region is probably causing the specific activation, but LRP and GBP did not. Considering that the orange color is the shared underlying visual information across the activated region, and our method together with Deconv consistently found the orange color in all the real and the synthetic images, it might suggest that the specific kernel does specifically respond to the orange color information and both our method and Deconv correctly interpret the activation of the kernel.

\begin{figure*}[tb]
		\centering
		\includegraphics[width=0.95\textwidth]{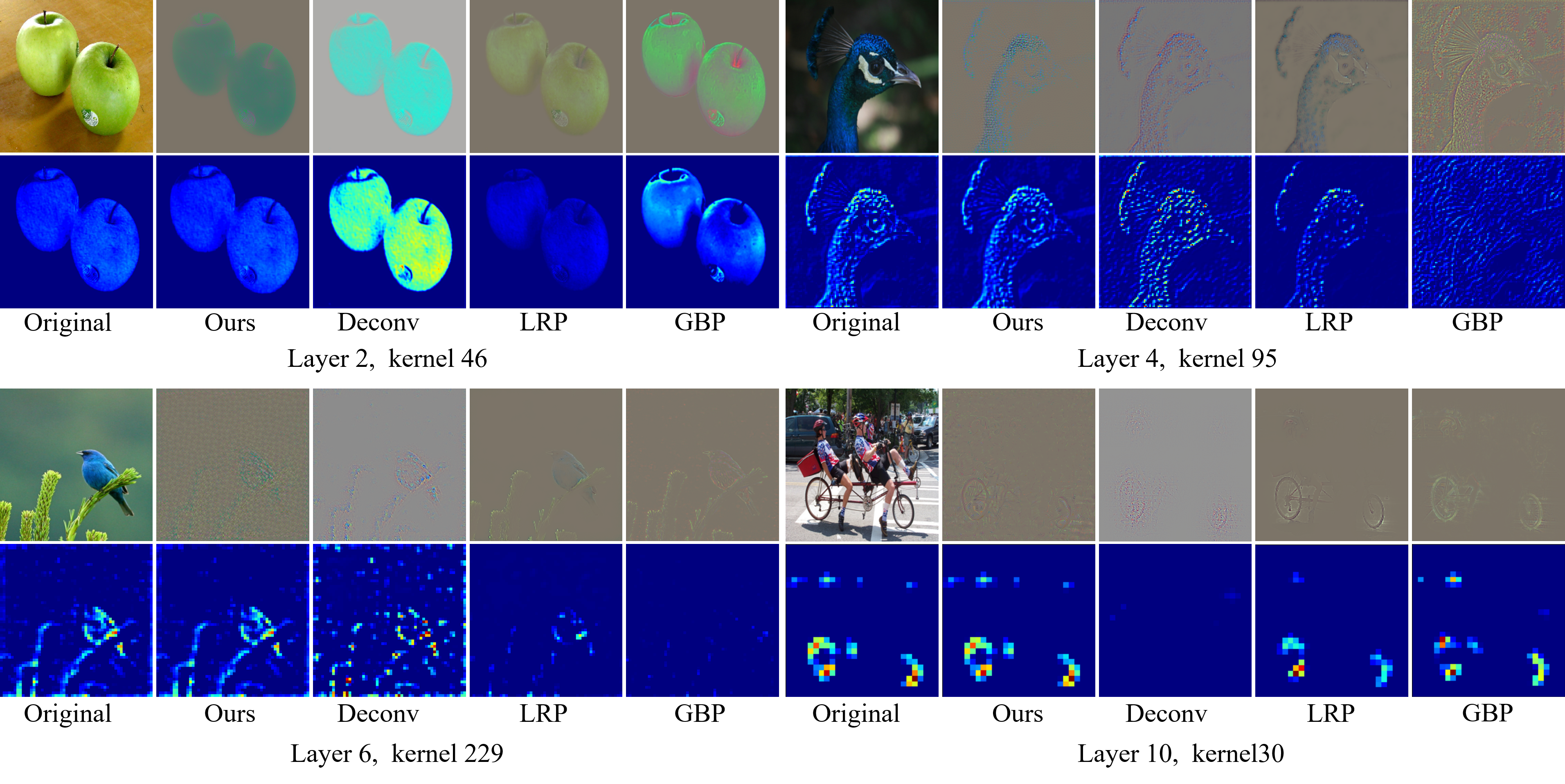}
		\caption{Optimized inputs (first and third row) and corresponding feature map activation (second and fourth row) of specific kernels at different convolutional layers. Our method can better preserve the activation of specific kernels at various layers.} 	\label{fig:comparison_layer}
	\end{figure*}

Figure~\ref{fig:comparison_layer} lists the optimized inputs and corresponding feature map activation of specific kernels at different convolutional layers. Compared to kernels at lower layers, interpretation of kernels at higher layer become more challenging. But one consistent observation is that the regions in the optimized inputs by all the methods are largely overlapped, indicating that all the methods are working in localizing the relevant regions for activation of specific kernels. However, only the optimized inputs from our method lead to the very similar feature map activation compared to that of the original inputs (second and last rows), and meanwhile better suppress the activation of other kernels at the same layer (not shown for simplicity). This again suggests that the visual information found by the baselines just partly contributes to the activation of specific kernels, while our method may find more complete and specific visual information for the activation of specific kernels.

	\begin{figure}[tb]
		\centering
		\includegraphics[width=0.47\textwidth]{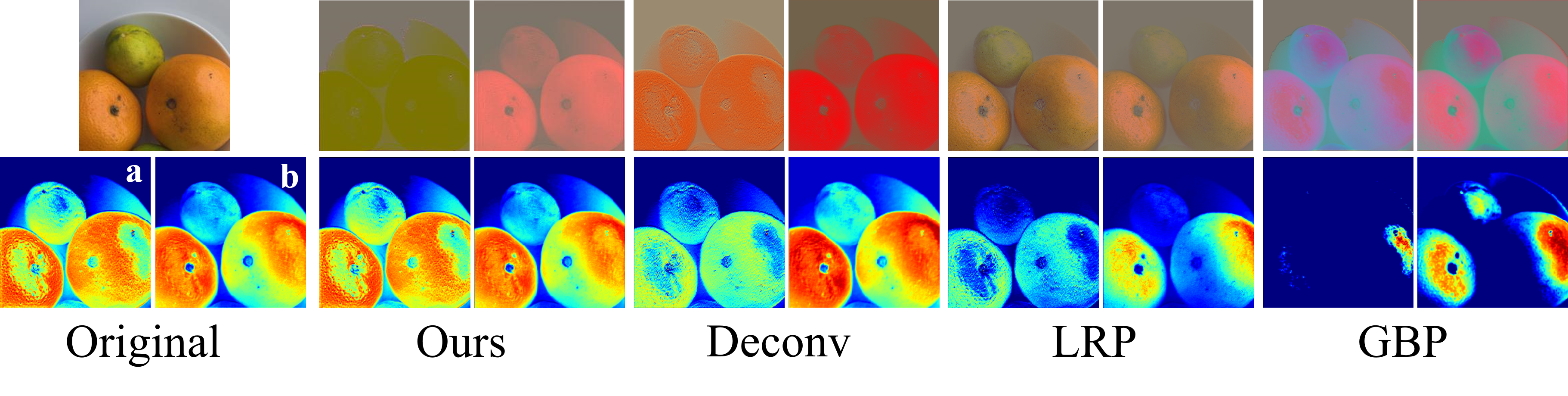}
		\caption{A demonstrative example of two kernels having very similar activation. First row: original and optimized inputs by different methods. Second row: feature map activation of two kernels with corresponding inputs in the first row.}
    	\label{fig:same_region}
	\end{figure}
	
    \begin{figure}[thb]
		\centering
		\includegraphics[width=0.47\textwidth]{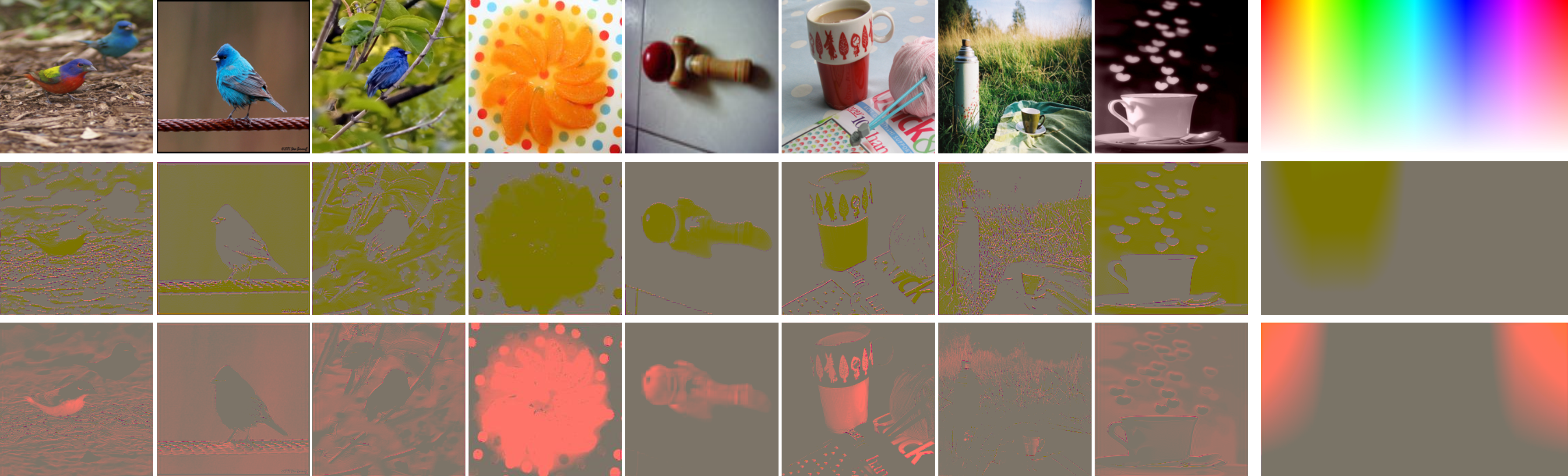}
		\caption{More interpretation results for the two kernels in Figure~\ref{fig:same_region}. Consistent visual information was observed in the optimized inputs for each kernel.}
		\label{fig:same_region_more}
	\end{figure}

Also probably due to simultaneous preservation of a specific kernel and suppression of other kernels, our method can more easily discriminate two kernels particularly when the two kernels respond to the same region of an image. Figure~\ref{fig:same_region} demonstrates an example of two kernels (`a' and `b') having almost the same feature activation (second row, first two) for the input image (first row, first). The optimized inputs (first row) for each of the two kernels were obtained respectively by our method and the baselines. Our method found two complete different visual information respectively for the activation of the two kernels, while the baselines found more or less similar visual information for the two kernels. In addition, from the corresponding feature map activation (second row), it can be seen that the optimized inputs by our method result in very similar activation to the original ones, while the estimated inputs by other methods contribute to only part of the original activation for one or both kernels. 
To further check whether the two types of visual information found by our method are really associated with the two kernels, multiple different original input images with various visual content were provided and analysed by our method (Figure~\ref{fig:same_region_more}). It shows that the optimized inputs for each of the two kernels by our method contain consistent visual information inside (Figure~\ref{fig:same_region_more}, last two rows). In particular, the optimized inputs for the synthetic color image (Figure~\ref{fig:same_region_more}, last column) contain the color information which is consistent  not only with the optimized inputs from other images but also with the color information in the original input. All these consistencies suggest that the two kernels are probably responding specifically to the two different color information which is  found only by our method.

\begin{figure}[!tbh]
		\centering
		\includegraphics[width=0.47\textwidth]{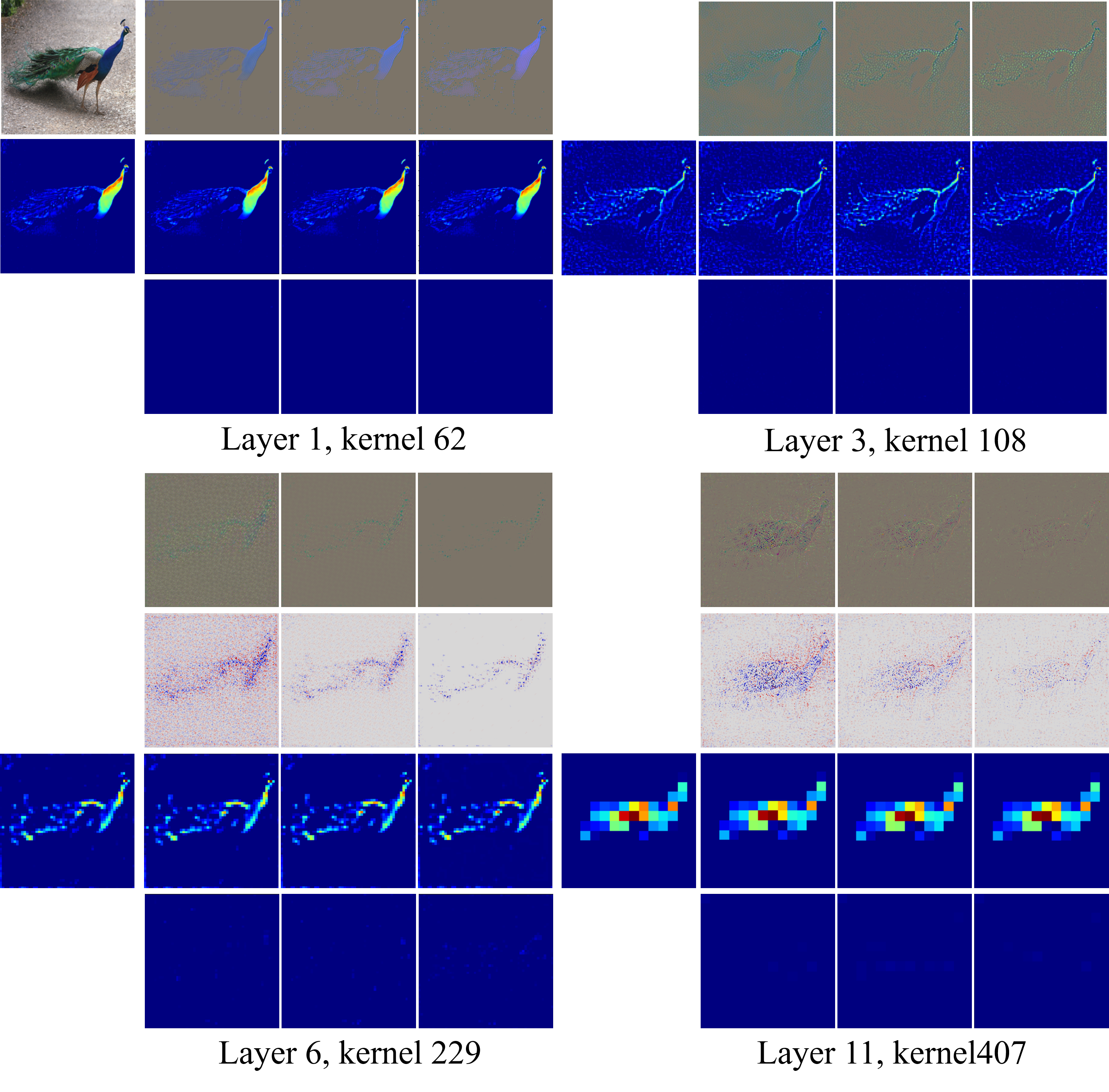}
		\caption{Multiple optimized inputs for the activation of each specific kernel obtained by varying the coefficient $\beta$ of the third loss term in our method. For the top two groups (except for the original input on the top left), first row: optimized inputs; second row: the original feature map and feature maps corresponding to the optimized inputs; third row: the difference between the original feature map and the feature map of each optimized input. For bottom two groups, one additional row (second row) was generated by linearly contrasting pixel values from the first row for clearer visualization. Each group is for one specific kernel at a different convolutional layer. 
	}
    	\label{fig:multisolution}
	\end{figure}

Although our method can help better interpret the activation of specific kernels than the baseline methods, the interpretation is probably still not comprehensive. In particular, by varying the coefficients (e.g., $\beta$) in the proposed loss function (Formula~\ref{eq:loss}), multiple optimized inputs can be obtained for the same original input and one specific kernel. As shown in Figure~\ref{fig:multisolution}, the multiple optimized inputs result in almost the same feature map activation as the original one (Figure~\ref{fig:multisolution}, last two rows for each group). While the multiple optimized inputs for each specific kernel contain similar information, the clear visual differences between them indicate that multiple input solutions exist for the same feature activation. Such differences between the solutions suggest that a specific activation of one kernel may be robust to certain variations in the input image. While the robustness is a desirable property for the classifier model to recognize objects which may appear more or less different in different images, it also causes the challenge of specific interpretation of the kernel activation (e.g., 'Is it the whole or part of the specific visual information in the input that contributes to the specific activation of the kernel?'). We leave this challenge as an open question for future study.

    \begin{table}[tb]
		\centering
		\footnotesize
        \caption{Ablation study of the suppression loss and the regularization loss. A tick means that the corresponding loss term was used for input optimization.}
		\label{tab:ablation_study}
        \vspace{0.5em}
        \begin{tabular}{ccccc}
        	\toprule
        	$\mathcal{L}_p$ & $\mathcal{L}_s$ & $\mathcal{L}_r$ & SSIM$\uparrow$(S.) & MSE$\downarrow$(R.) \\ 				
        	\midrule
        	\checkmark & \checkmark & & $0.82\pm 0.17$ & $0.34\pm 0.74$\\
        	\checkmark & & \checkmark & $\boldsymbol{0.89}\pm 0.11$ & $2.66\pm 3.22$\\
        	{\checkmark} & {\checkmark} & {\checkmark} & $0.84\pm 0.15$  & $\boldsymbol{0.25}\pm 0.03$ \\
        	\bottomrule
        \end{tabular}
	\end{table}

\subsection{Ablation study}
In addition, to further explore the effect of the suppression loss $\mathcal{L}_s$ and the regularization loss $\mathcal{L}_r$, ablation study was performed by removing one of them during input optimization. As table~\ref{tab:ablation_study} shows, the removal of the 
regularization loss $\mathcal{L}_r$ (first row) causes relatively lower SSIM and higher MSE, suggesting that the responses to original inputs were not very well preserved for kernels of interest and suppressed for the other kernels. When removing the suppression loss $\mathcal{L}_s$ only (second row), it is reasonable that the responses to original inputs were more likely preserved for kernels of interest (as demonstrated by higher SSIM) because there is less force to modify the original input. However in this case, the MSE becomes much higher due to the lack of force to suppress the responses of other kernels. 
In comparison, the combination of these three loss terms (third row) can achieve the best balance on suppressing irrelevant information for other kernels (lowest MSE) while preserving the information for the selected kernel (relatively high SSIM).

\section{Conclusion}
In this study, a simple yet effective optimization method is proposed to interpret the activation of any kernel of interest in a CNN model. By preserving the activation of the specific kernel and suppressing the activation of other kernels at the same layer, the proposed method can find visual information in the input image which is more specific to the activation of the kernel of interest. With consistent observation of the same type of visual information from the optimized inputs for the activation of the same kernel with multiple images, we could interpret that the kernel is specifically responding to such visual information. Quantitative and qualitative evaluation supports that the proposed method may better interpret kernels in the CNN model than existing methods. Future work includes more comprehensive interpretation of kernels for various model structures and applications to real scenarios like intelligent disease diagnosis.

{\small
   \bibliographystyle{ieee_fullname}
   \bibliography{egbib}
}

\end{document}